\documentclass[12pt, a4paper]{article}
\usepackage[utf8]{inputenc}
\usepackage[english]{babel}

\usepackage{biblatex}
\usepackage{graphicx}
\usepackage{csquotes}

\usepackage[margin=1in]{geometry}
\usepackage{hyperref}
\usepackage{xcolor}
\usepackage{gensymb}
\usepackage{array}
\usepackage{multirow}
\usepackage{booktabs}
\usepackage{adjustbox}
\usepackage[flushleft]{threeparttable}
\usepackage{amsmath,amssymb}
\usepackage{capt-of}
\usepackage{afterpage}
\usepackage{nonfloat}
\usepackage{subcaption}
\usepackage{lineno}

\title{Iterative Human and Automated Identification of Wildlife Images}

\author{Zhongqi Miao$^{1,2,\dagger,}$\thanks{zhongqi.miao@berkeley.edu; wgetz@berkeley.edu} , Ziwei Liu$^{3,}$\thanks{Equal Contributions.} , Kaitlyn M. Gaynor$^{4}$, Meredith S. Palmer$^{5}$, \\ Stella X. Yu$^{1,2}$, Wayne M. Getz$^{1,6,}$\footnotemark[1]}

\date{
\footnotesize 
$^1$Dept. Env. Sci., Pol. \& Manag., UC Berkeley, CA, United States\\
$^2$International Comp. Sci. Inst., UC Berkeley, CA, United States\\
$^3$School of Comp. Sci. \& Eng., Nanyang Tech. Univ., Singapore\\
$^4$Nat. Cent. for Eco. Ana. \& Syn., UC Santa Barbara, CA, United States\\
$^5$Dept. of Eco. \& Evo. Bio., Princeton University, NJ, United States\\
$^6$Sch. Math. Sci., Univ. KwaZulu-Natal, South Africa\\
}

\begin{document}

\maketitle

\begin{abstract}
    
    Camera trapping is increasingly used to monitor wildlife, but this technology typically requires extensive data annotation. Recently, deep learning has significantly advanced automatic wildlife recognition. However, current methods are hampered by a dependence on large static data sets when wildlife data is intrinsically dynamic and involves long-tailed distributions. These two drawbacks can be overcome through a hybrid combination of machine learning and humans in the loop. Our proposed iterative human and automated identification approach is capable of learning from wildlife imagery data with a long-tailed distribution. Additionally, it includes self-updating learning that facilitates capturing the community dynamics of rapidly changing natural systems. Extensive experiments show that our approach can achieve a $\sim90\%$ accuracy employing only $\sim20\%$ of the human annotations of existing approaches. Our synergistic collaboration of humans and machines transforms deep learning from a relatively inefficient post-annotation tool to a collaborative on-going annotation tool that vastly relieves the burden of human annotation and enables efficient and constant model updates.   
    
\end{abstract}


In our rapidly-changing world, continuous monitoring of natural systems is essential to understand and mitigate the impacts of human activity on ecological processes~\cite{steenweg2017scaling, rich2017assessing, barnosky2011has}. Recent technological innovations now allow for rapid collection of ecological data across vast spatial and temporal scales.However, the resulting information deluge creates a bottleneck for researchers who must process the data at management-relevant timescales~\cite{ahumada2020wildlife}. Artificial Intelligence (AI) offers promising  solutions for rapid and high-accuracy data processing~\cite{NorouzzadehE5716, miao2019insights}. The dynamic nature of ecological systems, however, poses unique challenges when developing accurate algorithms~\cite{liu2019large, liu2020open}. To overcome these hurdles, we showcase how the integration of limited human labor into the machine learning workflow can greatly increase both efficiency and accuracy of data processing.

\section{Introduction}

\subsection{Long-term camera trapping}

We are currently experiencing rapid, human-driven loss of global biodiversity ~\cite{hautier2015anthropogenic, barlow2016anthropogenic, ripple2017conserving, dirzo2014defaunation}. To understand the complex patterns, drivers, and consequences of species declines and extinctions, ecologists increasingly employ emerging technology to assist with data collection and processing. Motion-activated remote cameras (henceforth ``camera traps") have emerged as a popular non-invasive tool for monitoring terrestrial vertebrate communities~\cite{o2010camera, burton2015wildlife, kays2020born}. Decreasing cost and increasing reliability have recently led to the application of camera traps for long-term, continuous deployment aiming to monitor entire wildlife communities across multiple seasons and years ~\cite{swanson2015snapshot, steenweg2017scaling, ahumada2011community, pardo:inpress-snapshot}. Compared with one-time or annual surveys, continuous monitoring reveals new insight into wildlife responses to local, regional, and global environmental changes and to conservation interventions. This scale of monitoring is particularly valuable for capturing responses to environmental perturbations as they occur~\cite{steenweg2017scaling, rich2017assessing}. The `Snapshot Serengeti' project (www.snapshotserengeti.org), which has operated continuously since 2010, is a flagship example of a long-term camera trap monitoring program. Over the last decade, this survey has gathered unprecedented longitudinal data that have significantly enhanced our understanding of the seasonal and inter-annual dynamics of the Serengeti ecosystem~\cite{swanson2015snapshot, anderson2016spatial, palmer2017dynamic}. Projects of this magnitude have become increasingly common across eastern and southern Africa~\cite{pardo:inpress-snapshot} and around the world~\cite{steenweg2017scaling}. 

The greatest logistical barrier to long-term monitoring with camera traps is the overwhelming amount of human labor needed to annotate thousands or millions of wildlife images for ecological analysis~\cite{palmer:inprep-reinstating, ahumada2020wildlife, swanson2015snapshot, NorouzzadehE5716}. This annotation bottleneck creates a considerable mismatch between the pace of data collection and data processing, significantly curtailing the usefulness of camera trap data for on-going conservation and monitoring efforts~\cite{ahumada2020wildlife}. For example, a relatively modest camera trap survey ($\sim$80 camera traps; \cite{steenweg2017scaling}) captures millions of images a year. We estimate that it would take a single trained expert around 200 full-time working days to annotate one million images. As such, hundreds of human annotators (e.g., experts, trained volunteers, and citizen scientists) are required to keep pace with image accumulation. This need is likely to grow exponentially over the coming decades as more monitoring sites are set up. While only one or two experts are needed to validate each wildlife image, it is common practice that multiple (5-20) volunteers or citizen scientists look at each image in order to produce a high-accuracy "consensus" classification ($\sim$97\% accurate compared to expert IDs;~\cite{swanson2015snapshot}). This duplication of effort needed to generate accurate results using volunteers further perpetuates the classification bottleneck.

\subsection{Automatic image recognition systems}

The use of deep learning (a subset of AI technology) to automatically identify animals in camera trap images has recently drawn considerable attention from the ecological community. Currently, trained deep learning algorithms can classify a million images in a single day running on a desktop computer, a significant advancement over the months of effort required for human annotators to accomplish the same task~\cite{NorouzzadehE5716,tabak2019machine,whytock2021robust}.

There exist several attempts to develop robust camera trap recognition methods for real-world deployment, either tackling the distribution shift (in species numbers and locations) with transfer learning~\cite{beery2018recognition,tabak2020improving, shahinfar2020many}, or addressing the new species emergence with active learning~\cite{norouzzadeh2019deep,willi2019identifying,schneider2020three}. However, before it becomes feasible to rely on deep learning to handle the mass of image data from large-scale, long-term camera trap projects, two major impediments must be overcome: 1) accounting for temporal changes in species composition at study sites due to migration, invasion, re-introduction, and extinction and 2) handling the long-tailed distribution of records across species (i.e., extreme imbalance in the number of images of different species, Fig.~\ref{fig:totaldist} in the Supplementary Method section). As discussed below, these issues limit the ability of current AI to accurately recognize species that are of significant interest to conservation practitioners.

\subsubsection*{Changing species composition}

A novel challenge for long-term surveys is that new species may be detected on cameras in subsequent seasons or years, either because the species are rare and undetected in the previous survey periods~\cite{kays2020empirical}, or because they are new to the system. Additionally, the species composition of ecological systems naturally varies through time through the process of succession~\cite{prach2011four}. Novel species are often of particular conservation concern, as they may represent recolonizing populations~\cite{mech2019gray}, reintroduced animals~\cite{taylor2017reintroduction, palmer:inprep-reinstating}, or harmful invasive species~\cite{clavero2005invasive, caravaggi2016invasive}.  

In conventional deep learning, researchers focus on the performance of existing testing data while ignoring the potential for future changes in data composition~\cite{arjovsky2019invariant}. In other words, deep learning models typically require data sets to be fixed in number of categories (i.e., static), while in reality, long-term camera trap data sets are not constrained to certain numbers of species (i.e., dynamic).

Fine-tuning models through transfer learning is currently the best solution when new species populate a study area~\cite{yosinski2014transferable}. However, this processes requires full annotation of newly-collected data sets, requiring a considerable amount of new human effort. This defeats the purpose of deep learning to reduce manual labor for long-term camera trap monitoring.

 
\subsubsection*{Data from wildlife communities are long-tailed}

Wildlife communities typically contain many individuals of several common species and few individuals of many rare species, resulting in camera trap data with a long-tailed distribution. For example, in the data set used for this project from Gorongosa National Park, Mozambique, $\sim$50K images ($>60\%$ of animal images) are of baboon, warthog, and waterbuck, while only 22 images are of pangolin (a rare and protected species). This imbalance creates performance inconsistencies because deep learning success is derived from balanced training data sets (e.g., ImageNet \cite{deng2009imagenet}). For the Gorongosa data set, a traditional deep learning approach resulted in only 60\% accuracy for a category with only 41 images (serval) versus 88.8\% performance for a species with 17,938 images (waterbuck). This is a major issue because animals of particular conservation concern are typically rare~\cite{pimm2014biodiversity}, producing less images and therefore worse classification accuracy than common species. If such species are always misclassified, AI's practical benefits are limited. 

\subsection{An iteratively updating recognition system}

To overcome these two major issues of 1) changing species community composition and 2) long-tailed species distributions, we designed a deep learning recognition framework that is updated iteratively using limited human intervention. 
Human annotation is needed whenever images of species novel to the AI model appear in the data. Our goal, therefore, becomes to minimize the need for human intervention as much as possible by applying human annotation solely on difficult images or novel species, while maximizing the recognition performance/accuracy of each model update procedure (i.e., update efficiency). 

Traditionally, a deep learning model is applied to new batches of unannotated data collected during each time period to predict species classes. In our approach, we actively flag images that our model predicts with low-confidence as novel or unknown species. These low-confidence predictions are then selected for human annotation while high-confidence predictions are accepted as accurate and used as pseudo-labels for future model updates. Then, the model is updated (i.e., retrained) based on both human annotations and pseudo-labels. To accommodate changing species communities, this procedure of active annotation and model update repeats each time new data are added to the collection (Fig.~\ref{fig:main}). In terms of long-tailed distribution, we use Open Long-tailed Recognition (OLTR)\cite{liu2019large} method to balance the learning between abundant and scarce species. This component can reduce the number of predictions with low-confidence from scarce species.

As a case study, we trained a model on a camera trap data set collected from Gorongosa National Park, Mozambique (see the Supplementary Method section for details) using this new method and produced significantly improved model update efficiency over traditional transfer learning approaches. Specifically, more than 80\% human effort is saved on annotating new data without sacrificing classification performance using our approach.

The dynamic nature of our algorithm maximizes learning and recognition efficiency by taking the best from both humans and machines within a synergestic collaboration. To the best of our knowledge, our model is the first framework that can be practically deployed for long-term camera trap monitoring studies. 




\section{Iterative Human and Automated Identification}

\subsection{Algorithm overview}
Our approach has two major components: 1) active selection with humans in the loop, and 2) model update using active data annotations. At each time period when new data are collected, categories of images are predicted by deep learning models trained from previous periods with corresponding confidence levels. The model actively picks out low-confidence predictions for human annotation, while we accept high-confidence predictions without further human verification as accurate. These predictions are used as pseudo-labels that are included in the final data set for further model updates or ecological analyses. Next, the model is updated (retrained) using both pseudo-labels and the newly acquired human annotations (see the Supplementary Method section for implementation details). 

After updating the model, we evaluated model-update efficiency and sensitivity to novel categories on a validation set. Specifically, we examined: 1)overall validation accuracy of each category after the update (i.e., update performance); 2) percentage of high-confidence predictions on validation (i.e., saved human effort for annotation); 3) accuracy of high-confidence predictions; and 4) percentage of novel categories that are detected as low-confidence predictions (i.e., sensitivity to novelty). The optimization of the algorithm aims to minimize human efforts (i.e., to maximize high-confidence percentage) and to maximize model update performance and high-confidence accuracy.

\subsection{Data Specifications}

\subsubsection*{Data Categories}
We manually identified at total of 55 categories (i.e., species) in our data, including non-animal categories, such as ``ghost'' (i.e., misfired images lacking animals), ``setup'' (i.e., images with human setting up the cameras), and ``fire.'' There were 630544 images in total. The full list of these categories is in Fig.~\ref{fig:totaldist} in the Supplementary Method section along with the number of images associated with each category. Some ``vague'' categories that human annotators were unable to accurately label because of the varying quality of camera trap images were also present, such as ``unknown antelope'' and ``unknown bird''. 

\subsubsection*{Two groups of training and validation sets}
To ensure sufficient training and validation data, we initially identified 41 of the most abundant categories in our camera trap data set. The remaining 14 of the 55 categories were all tagged as ``unknown'' and used to improve and validate the model's sensitivity to novel and difficult samples. We randomly split the 41 categories (by trigger events) into \textbf{two groups of training and validation sets} (26 categories in the first group of data and 41 in the second group) to mimic periodical data collection from two sequential time periods. Detailed training and validation split information can be found in the Supplementary Method section.

\subsection{Detailed pipeline for experiments}

For experimental purposes, we separated our identification pipeline into two steps representing two time periods of data collection and the two groups of data curated in this project (Fig.~\ref{fig:workflow}). The evaluation is focused on the second period when model update occurs. There are three major technical components in the framework: 1) energy-based loss~\cite{liu2020energy} that improves the sensitivity to possible novel and difficult samples for active selection; 2) a pseudo-label-based semi-supervised procedure~\cite{lee2013pseudo} for efficient model update from limited human annotations; and 3) open long-tailed recognition (OLTR)~\cite{liu2019large} that balances the learning of long-tailed distribution.

\subsubsection*{Period 1}
In the first period, we pre-trained an off-the-shelf model (ResNet-50 model~\cite{he2016deep}) using the first group of data. 
After training, we adopted the energy-based loss~\cite{liu2020energy} and data from the 14 ``left-out'' categories to fine-tune the classifier so it is more sensitive to novel and difficult samples. 

\subsubsection*{Period 2}
In the second period, we first used the fine-tuned model from Period 1 to produce high- and low-confidence predictions from group 2 training data, which were considered to be ``newly collected''. The confidence was calculated based on the \textit{Helmholtz free energy} (see the Supplementary Method section for details) of each prediction~\cite{liu2020energy}. Novel and difficult samples were distinguished using a preset energy threshold. Then, low-confidence predictions were annotated by humans while high-confidence predictions were accepted as pseudo-labels.

To update the model, we applied semi-supervised learning and OLTR, using both human annotations and pseudo-labels. Pseudo-label-based semi-supervised approaches iteratively update both the model and pseudo-labels until the best performance on the validation sets is achieved~\cite{lee2013pseudo}. The use of pseudo-labels also enables the model to learn from the whole data set instead of human annotated data only. On the other hand, OLTR approaches balance the learning between abundant and scarce categories through an embedding space memory-based mechanism, where embedding memories of abundant categories are utilized to enhance the distinguishiability of scarce categories that do not have enough samples to otherwise provide discriminative features~\cite{liu2019large}. (See the Supplementary Method section for details of these methods.) 

After the model is updated, the training sample from the 14 ``left-out'' categories were added to fine-tune the model's sensitivity to novel and difficult samples using energy-based loss as in Period 1. 

\subsubsection*{Future Periods}
Because the framework is designed to aid long-term data collection and monitoring projects, the framework does not stop at Period 2. As time progresses, new data are collected. Users simply have to repeat the steps in Period 2 to pick out and annotate difficult/novel samples to update the model. In addition, since the framework is fully modular, when new techniques are developed, parts of the framework can be easily replaced for better performance. For example, if there are better methods for novel-category-detection, energy-based loss and confidence calculation can be replaced with no effect on the conceptual framework.


\section{Results}

\subsubsection*{Period 1}
In the first period, the model achieved an 81.2\% average class accuracy on the validation set of group 1, 79.5\% of the images predictions were high-confidence, and of these predictions, the accuracy was 91.1\% (Table~\ref{tab:overall_performance} and Table~\ref{tab:active_performance}). In terms of novel categories, in the validation phase, the model successfully detected 90.1\% of the novel samples belonging to the 14 categories that were ``left-out'' of the training phase. In other words, 90.1\% of the novel samples were predicted with low-confidence. In contrast, direct Softmax confidence (the most conventional way of calculating prediction confidence~\cite{hinton2015distilling}) achieved a similar high-confidence accuracy as our model (91.5\%), but only detected 59.3\% novel samples.


\subsubsection*{Period 2}
On group 2 training data, the model pretrained from Period 1  predicted 78.7\% images with high-confidence where the accuracy was 92.4\%. 75.7\% of the new categories in group 2 training data were detected as low-confidence predictions (Table~\ref{tab:active_performance}). As high-confidence predictions are trusted, 78.7\% human effort was saved to annotate group 2 training data because high-confidence predictions were accepted as accurate in our framework.

To update the model, group 2 training data that had been predicted with low-confidence were checked by human experts and provided with manual annotations, and high-confidence samples were assigned model-predicted pseudo-labels. Overall, on the validation set of group 2, the model updated on both human annotations and pseudo-labels had an average class accuracy of 77.2\% over the 41 categories. Compared to our method without human annotation (69.2\%; second to the last row in Table~\ref{tab:overall_performance}), there was an 8\% improvement. The model had a 72.3\% high-confidence predictions at a 90.2\% accuracy in the high-confidence predictions of the validation set (see Table~\ref{tab:overall_performance}, Table~\ref{tab:active_performance}, and Table~\ref{tab:class_perf} for detailed per-category performances). In addition, it had a 82.6\% novel sample detection rate (i.e., flaged as low-confidence predictions) from the validation data of the 14 ``left-out'' categories (see the last column of Table~\ref{tab:active_performance}). 

\subsubsection*{Comparison with traditional transfer learning}
Our model was significantly more data efficient (i.e., less data required for the same performance) than traditional transfer learning methods in several respects. Compared to traditional transfer learning, which used full human annotations of group 2 training data, our method only involved human annotation of 21.3\% of the group 2 samples. Even with less human annotation, our method still achieved better overall class average accuracy (77.2\% vs. 75.8\% for traditional transfer learning; Table~\ref{tab:overall_performance},\ref{tab:active_performance}). Our model also performed better than direct transfer learning for classifying the 15 new categories from Group 2 (with an average of 4.2\% accuracy improvement; Table~\ref{tab:class_perf}).

\subsubsection*{Practical deployment}
Our new framework showcases the powerful potential of deep learning for long-term ecological application while employing a novel practical approach that greatly reduces the manual annotation burden. To validate the practical benefits, we deployed the model to classify a new set of data gathered from the same camera trap monitoring sites (Gorongosa National Park, Mozambique) after group 1 and 2 data sets were collected (see the Supplementary Method section for details). The new data set are unannotated, unanalyzed, and contained 623,333 images in total. Images were predicted with the same active selection procedure and 78.7\% of the predictions were considered high-confidence. Thus only 21.3\% of these newly-collected data required human annotation (or 78.7\% of the human effort, and ultimately annotation cost, was saved). 

To validate the robustness of model performance, two experts (KMG and MSP) confirmed the accuracy of 1000 randomly-selected high-confidence predictions (i.e., those that were accepted as accurate). Our model predictions are 88.6\% accurate with respect to expert classifications. Statistically, $\sim$88\% automatic accuracy is already sufficient to help alleviate the data bottleneck encountered in typical camera trap monitoring projects compared to expert accuracy.

In terms of future model update, the model can be further updated and validated on the new data set using the same procedure as Period 2, where a new validation set can be created using a mix of previous validation sets (validation of groups 1 \& 2) and the newly acquired human annotations. In addition, the same random verification by human experts on high-confidence predictions can be applied to avoid performance corruptions (i.e., increased misclassifications in high-confidence predictions).

\subsubsection*{Invasive and recolonizing species}
One of the significant advances made by our framework is the ability to flag new or rare species that may have particular conservation importance. Our new data set contained two novel species (leopard and African wild dog) to test the model's sensitivity to novel categories. The former naturally re-colonized the study area while the latter was re-introduced as a part of on-going conservation efforts. There were 24 and 5 images for African wild dogs and leopards respectively. The model successfully detected 20 (83.3\%) African wild dog and 4 (80.0\%) leopard images, demonstrating its capacity to recognize important novel species in continuous monitoring periods.

\section{Discussion}

\subsubsection*{Failure cases}
Two types of failures occur in our framework: 1) low-confidence predictions that are not novel species, and 2) high-confidence predictions that differ from human-supplied annotations. 

First, there are several ways in which our model was unable to accurately identify samples from known species with high-confidence (Fig.~\ref{fig:failures}a). A common reason for low-confidence predictions was difficulty distinguishing animals from the background. For example, Fig.~\ref{fig:failures}a.i depicts an antelope obscured by darkness at night, making it difficult for the model to confidently classify. However, rather than making a misclassifications as would occur in traditional AI approaches~\cite{he2016deep}, our model considers the low accuracy of the prediction and flags the image for review or similar. In our approach, these difficult samples are flagged as low-confidence predictions for further human evaluation (annotation) rather than assigned random labels---a practice which can potentially bias further data analysis and inference.

In the second type of model failure, images predicted with high-confidence differ from the original annotations (Fig.~\ref{fig:failures}b). We note that these images were originally classified by volunteers who were trained but may not have correctly annotated all samples as accurately as wildlife experts. Surprisingly, most of the confident predictions are proven to be correct after re-evaluation by human experts (KMG and MSP). For example, Fig.~\ref{fig:failures}b.iv was originally labeled as a warthog, although there is no warthog present. However, there is a vervet monkey, in the lower left of the frame that was missed by the human classifiers. The model not only detects the previously unobserved animal but also correctly identifies the species.

Thus, these ``failures" actually demonstrate the robustness and flexibility of our framework. As both human annotations and machine predictions can be wrong, a mutual interaction between human and machine can benefit long-term performance of the recognition system. For example, picking out low-confidence samples like the ones in Fig.~\ref{fig:failures}b prevents producing low quality predictions that can cause bias in camera trap analyses. Further, applying validated human annotations on these samples can help improve the identification capacity of the model as it needs to recognize more difficult samples during model updates. On the other hand, when the model is highly confident, it can be more accurate than average human annotators, as evidenced by the examples given in Fig.~\ref{fig:failures}b.ii, iv, and v). In other words, some of the human mistakes are prevented, such that the annotation quality for future model update and camera trap analyses are improved. On the other hand, as we acknowledge that in some of cases, the model will make high-confidence classifications, we can apply periodical random verification by human experts on high-confidence predictions (similar to what we did in Practical Deployment section) to ensure that these errors do not propagate through repeated training.

\subsubsection*{The need for humans in the loop}
Our framework demonstrates the unique merit of combining machine intelligence and human intelligence. As Fig.~\ref{fig:failures}c illustrates, machine intelligence, when trained on large data sets to distill visual associations and class similarities, can quickly match visual patterns with high confidence~\cite{deng2009imagenet}. Human intelligence, on the other hand, excels at being able to recognize fragmented samples based on prior experience, context clues, and additional knowledge. 
Increasingly, we are moving towards applying computer vision systems to real-world scenarios, with unknown classes~\cite{liu2019large}, unknown domains~\cite{liu2020open}, and constantly-updating environments. It is therefore crucial to develop effective algorithms that can handle dynamic data streams.
Humans in the loop provides a natural and effective way to integrate the two types of perceptual ability (i.e., human \& machine), resulting in a synergism that improves the efficiency and of the overall recognition system.

\subsubsection*{Extensions and future directions}
Our framework is fully modular and can be easily upgraded with more sophisticated model designs. For example, models with deeper networks can be employed for better classification generalization, more sophisticated semi-supervised training protocols can be adopted for better learning from pseudo-labels, and better novelty detection techniques can be used for better active selection. 

Future directions include extending our framework to handle multi-label and multi-domain scenarios. The current approach was developed for single-label recognition (i.e., each image only represents one single species). In real-world camera trap setups, it would be desirable to recognize multiple species within the same view. Further, our framework is expected to be deployed in diverse locations with different landscapes. Therefore, our methodology can be more scalable with the ability to handle multiple environmental domains than existing methodologies. In addition, our method will be incorporated in a user-friendly interface, such that users without knowledge of Python can use it.

\section{Acknowledgement}

Thanks to T. Gu, A. Ke, H. Rosen, A. Wu, C. Jurgensen, E. Lai, M. Levy, and E. Silverberg for annotating the images used in this study, and to everyone else involved in this project. Data collection was supported by J. Brashares and through grants to KMG from HHMI BioInteractive, the Rufford Foundation, Idea Wild, the Explorers Club, and the UC Berkeley Center for African Studies. We are grateful for the support of Gorongosa National Park, especially M. Stalmans in permitting and facilitating this research. ZL is supported by NTU NAP. KMG is supported by Schmidt Science Fellows in partnership with the Rhodes Trust, and the National Center for Ecological Analysis and Synthesis Director's Postdoctoral Fellowship. MSP is funded on National Science Foundation grant, PRFB \#1810586.

\section{Author contributions statement}

This study was conceived by ZM, ZL, KMG, and MSP. The methods were designed by ZM and ZL. Code was written by ZM, and the computations were undertaken by ZM with help from ZL. The main text was drafted by ZM and ZL with contributions, editing, and comments from all authors. KMG and MSP collected all data and oversaw annotation. ZM created all figures and tables in consultation with WMG, ZL, and SXY. ZM and ZL had equal contributions to this paper.

\section{Competing interests}
The author(s) declare no competing interests.

\clearpage

\begin{figure*}[!htb]
  \centering
  \includegraphics[width=\textwidth]{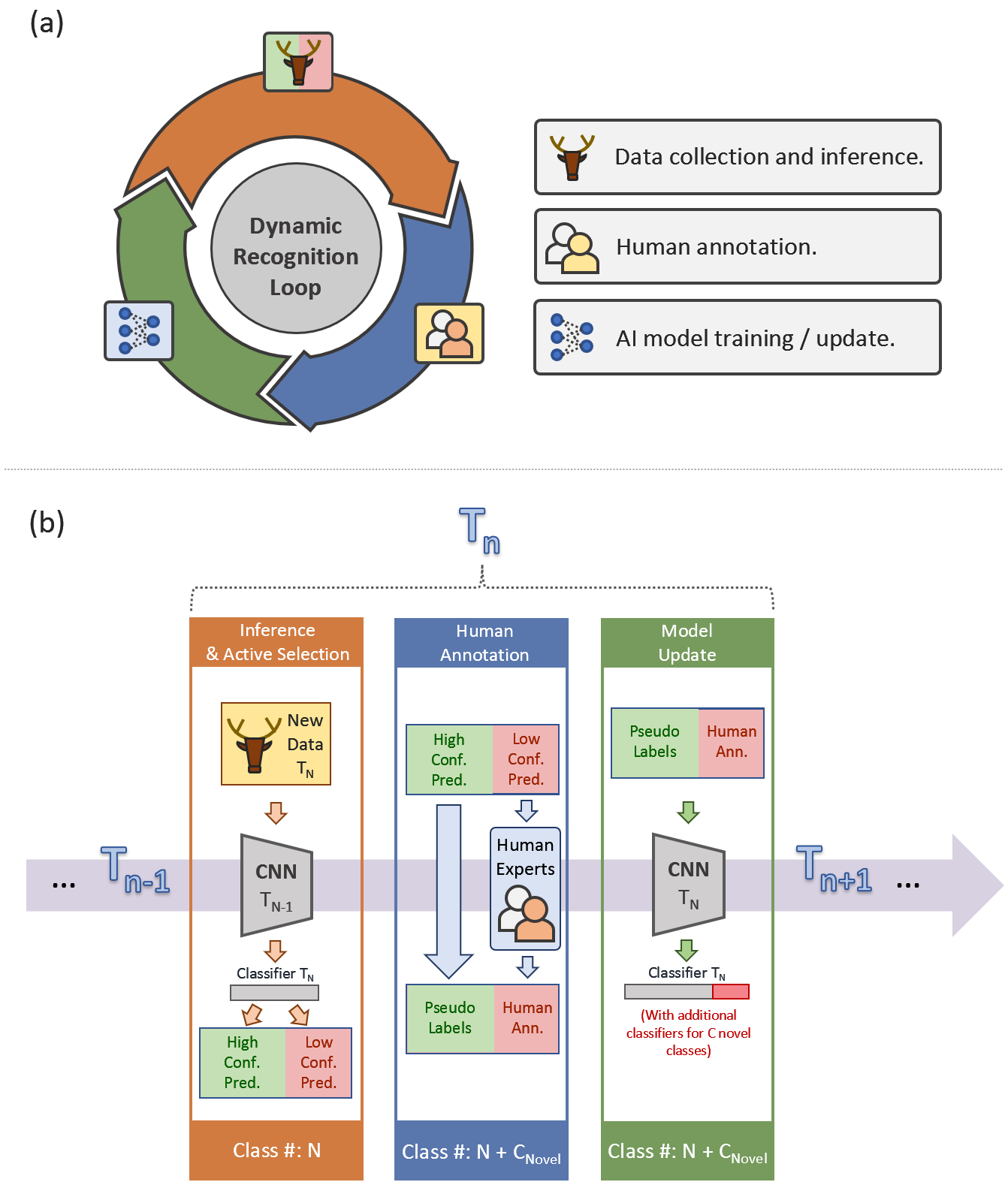}
  \vspace{-10pt}
\end{figure*}
\figcaption{\small \textbf{a) Dynamic recognition loop.} In real-world applications, machine learning models do not stop at one training stage. As data collection progresses over time, there is a continuous cycle of inference, annotation, and model updating. Every time a tranche of new data are added, pretrained models are applied to classify data. When there are novel and difficult samples, human annotation is required, and the model needs to be updated to reflect the newly added data. 
\textbf{b) The progression of a realistic animal classification system.} Even if the trained model has high accuracy for the previous validation sets, there may be a difference in the classes between previous validation sets and current inference data (e.g., there may be novel categories in the newly collected data that did not exist in previous training and validation sets). Models therefore need to be updated over time. Here, we present a more practical procedure that can both maximize the utility of modern image recognition methods and minimize the dependence on manual annotations for model updating. In this procedure, we incorporate an active learning technique that actively selects low-confidence predictions for further human annotation, while keeping highly-confident predictions as pseudo labels. Models are then updated according to both human annotations and pseudo labels. $^*$Symbols: $T$ is time step. $CNN$ is convolutional neural networks. $N$ is the total number of classes at time step $T_{n-1}$. $C_{Novel}$ is the number of novel classes at time step $T_n$. }  \label{fig:main}

\clearpage

\begin{table}[!htbp]
\centering
\caption{\textbf{classification performance comparisons on validation sets of periods 1\&2.}}
\label{tab:overall_performance}

\begin{threeparttable}
{\scriptsize
\begin{tabular}{
>{\arraybackslash}m{0.06\textwidth} |
>{\arraybackslash}m{0.5\textwidth} |
>{\centering\arraybackslash}m{0.12\textwidth} 
>{\centering\arraybackslash}m{0.18\textwidth} 
}

\toprule

\textbf{Periods} & \centering\textbf{Methods} & \textbf{Class Avg. Acc. (\%)} & \textbf{Class Avg. Acc. On New Classes. (\%)}\\
\hline
\centering\textbf{1} & \textbf{Off-the-shelf Model} & 81.2 &  - \\
\hline
\multirow{3}{0.06\textwidth}{\centering\textbf{2}} 
& \textbf{Traditional transfer learning w/ full human ann.} & 75.8 & 63.9 \\
& \textbf{Our framework w/out semi-supervision and OLTR} & 69.2 & 61.2 \\
& \textbf{Our framework (Semi-OLTR)} & \textcolor{red}{77.2} & \textcolor{red}{68.1} \\

\bottomrule

\end{tabular}

\begin{tablenotes}
\small
\item Red color means higher performance on the \textbf{same} inference set.
\item w/ : with.
\item ann : annotation.
\item Avg. : Average.
\item Acc. : Accuracy.
\end{tablenotes}
}
\end{threeparttable}
\end{table}

\begin{table}[!htbp]
\centering
\caption{\textbf{Active selection performances of Period 1\&2 with and without energy based function.}}
\label{tab:active_performance}

\resizebox{\textwidth}{!}{
\begin{threeparttable}
{\footnotesize
\begin{tabular}{
>{\arraybackslash}m{0.08\textwidth} |
>{\arraybackslash}m{0.17\textwidth} |
>{\arraybackslash}m{0.17\textwidth} |
>{\centering\arraybackslash}m{0.1\textwidth}
>{\centering\arraybackslash}m{0.1\textwidth}
>{\centering\arraybackslash}m{0.1\textwidth}
}

\toprule

\textbf{Periods} & \centering\textbf{Inference sets} & \centering\textbf{Confidence Metrics} & \textbf{High Conf. Ratio (\%)} & \textbf{High Conf. Acc. (\%)} & \textbf{Novel Detect Ratio (\%)} \\
\hline
\multirow{2}{0.08\textwidth}{\centering\textbf{1}} 
& \textbf{Group 1 Val.} & \textbf{Softmax} &  \textcolor{red}{80.9} & \textcolor{red}{91.5} & 59.3 \\
& \textbf{Group 1 Val.} & \textbf{Energy (Ours)} & 79.5 & 91.1 & \textcolor{red}{90.1} \\
\hline
\multirow{3}{0.08\textwidth}{\centering\textbf{2}} 
& \textbf{Group 2 Train} & \textbf{Energy (Ours)} & 78.7 & 92.4 & 75.7 \\
& \textbf{Group 2 Val.} & \textbf{Softmax} & 71.2 & 90.1 & 70.5 \\
& \textbf{Group 2 Val.} & \textbf{Energy (Ours)} & \textcolor{red}{72.2} & \textcolor{red}{90.2} & \textcolor{red}{82.6} \\

\bottomrule

\end{tabular}
\begin{tablenotes}
\small
\item Red color means higher performance on the \textbf{same} inference set.
\item Conf. : Confidence.
\item Acc. : Accuracy.
\end{tablenotes}
}
\end{threeparttable}
}

\end{table}

\clearpage

\begin{figure*}[!htb]
  \centering
  \includegraphics[width=\textwidth]{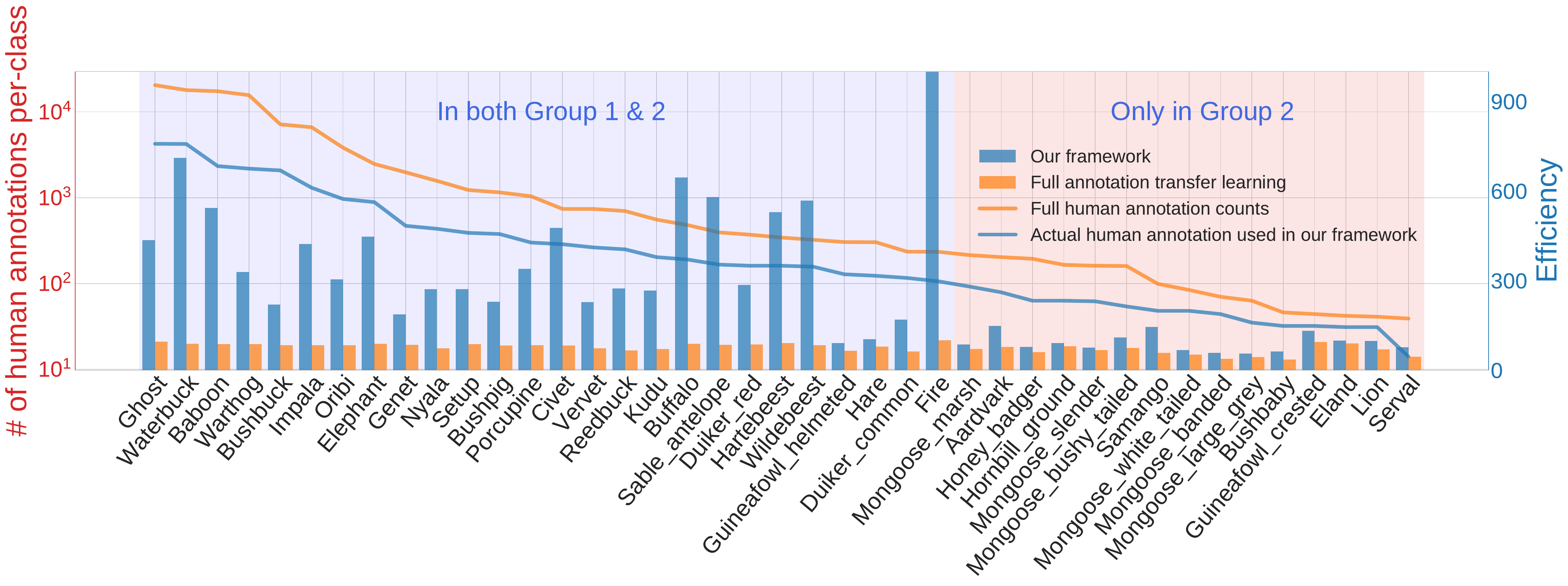}
  \caption{\textbf{\small Label efficiency comparison with transfer learning on Group 2 validation set (ordered with respect to training sample size).} To examine label efficiencies (a measure of accuracy given number of annotations) after we update our model in Period 2, we calculated validation accuracy over the percentage of used training annotations of each category. In other words, we define label efficiency: $\mbox{Efficiency}_{i} = \mbox{Validation Accuracy}_{i} / (\mbox{\# of training annotation}_{i} / \mbox{\# of full annotation}_{i})$ , where $i$ is the category index. The higher the value is, the more efficient the model is at learning corresponding categories, and the less training data are needed to achieve comparable if not better performance of full manual annotations. In the figure, we illustrate label efficiencies of all categories exist in Group 2 training and validation set. The blue bars represent our model's label efficiencies of each category. The orange bars represent baseline efficiencies for comparison, where full annotations were used with traditional transfer learning method (i.e., $\mbox{\# of training annotation}_{i} / \mbox{\# of full annotation}_{i} = 1$). The two blue and orange lines are annotation counts of each categories, where brown represents full annotations, and green represents actually used human annotations in our Period 2 model update procedure. For categories that exist in both the Group 1 \& 2 training sets (i.e., known categories; on the left, with a blue background), the efficiency is significantly higher than the baselines across all categories. For categories that only exist in Group 2 data sets (i.e., they were absent in the Group 1 training and validation set; novel categories; on the right, with an orange background), the model is designed to use as much training data as possible because of the novelty of these categories. In other words, $\mbox{\# of training annotation}_{i} / \mbox{\# of full annotation}_{i}$ of these categories are close to 1. Our model still has relatively higher efficiency than the full annotation transfer learning model across all the novel categories because our model had higher validation accuracy with similar amount of training annotations.}
  \label{fig:efficiency}
  \vspace{-10pt}
\end{figure*}

\clearpage

\begin{figure*}[!htb]
  \centering
  \includegraphics[width=\textwidth]{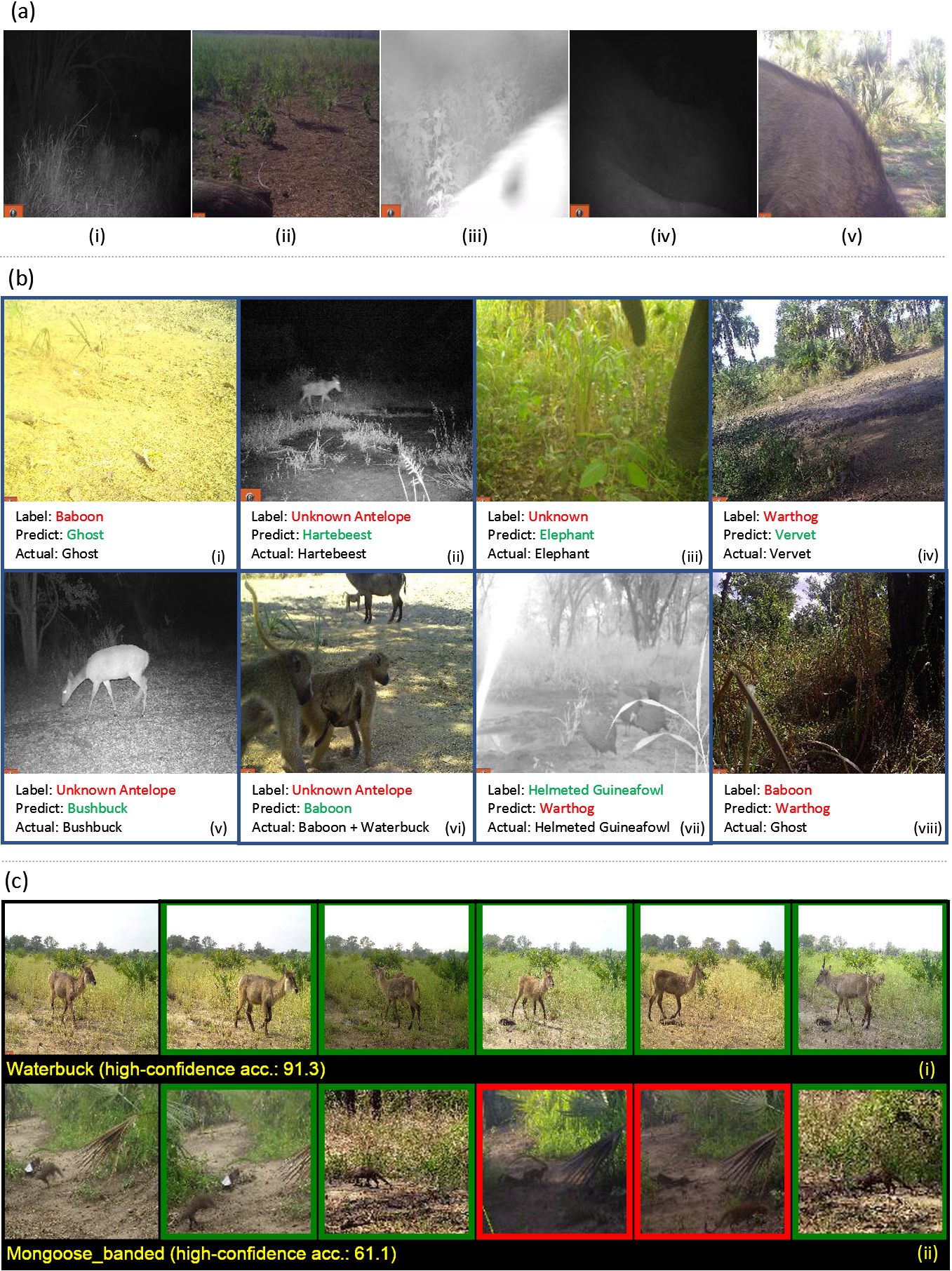}
  \vspace{-10pt}
\end{figure*}
\figcaption{\small \textbf{(a) Examples of low-confidence predictions.} In most of the cases, the model has low confidence on images with distorted, partially visible (panel ii$\sim$v), or obscured animals (panel i). It can be incredibly difficult, if not impossible, for both humans or machines to accurately identify the animal species. \textbf{(b) Examples of high-confidence predictions that did not match the original annotations.} Many high-confidence predictions that were flagged as incorrect based on validation labels (provided by students and citizen scientists) were in fact correct upon closer inspection by wildlife experts (KMG and MSP). For example, Panel (i), an empty image and originally mislabeled as baboon, was correctly classified by our method as empty. In panel (ii), although the animal is distant from the camera in a dark environment, the model successfully identifies hartebeest, while the human-supplied label is ``unknown antelope''. In panel (iii), the model successfully identifies the elephant only based on the trunk and leg, while human volunteers originally classified the image as "unknown". In panel (iv), a vervet monkey is correctly detected and classified in an image originally (incorrectly) labeled as warthog by human annotators. Panel (v) was originally classified as unknown by human annotators, but based on the body shape and white markings on the rear, the model can correctly recognized the animal as bushbuck. Panel (vi) is an example where multiple species are in the same scene. Although the model does not have capacity to deal with multi-species samples, as baboon is obviously the major component of this image, the prediction is reasonable. On the other hand, these examples above do not mean that the model always makes correct predictions when highly confident. Panels (vii) and (viii) are two typical examples where the model makes mistakes due to the obscured nature of these images. Red text indicates wrong, and green text indicates correct. \textbf{(c) Two examples of image retrieval based on feature space similarity.} Machine intelligence largely depends on visual similarity associations learned from large-scale data sets to classifies animal species. These two examples illustrate image retrieval based on the Euclidean distances of the feature vectors (i.e., outputs of the global average pooling layer of the ResNet model used in the project, which is of dimension 2048 in Euclidean space). For each anchor image (the leftmost image of each row), we show five closest (i.e., most similar) samples in terms of Euclidean distance within the validation set of Group 2. Green color means correct predictions, and red color means``wrong'' predictions (based on the original annotations). For example, in sequence (i), samples with similar visual appearances are usually from the same species (waterbuck). However, in sequence (ii), two most similar images (according to our model) to the banded mongoose anchor image are actually not banded mongoose but slender mongoose. The model misclassified these two samples based on their similarities to the other banded mongoose images.\label{fig:failures}}

\clearpage

\printbibliography

\clearpage

\appendix
\setcounter{figure}{0}
\setcounter{table}{0}
\renewcommand\thefigure{\Alph{section}.\arabic{figure}}
\renewcommand\thetable{\Alph{section}.\arabic{table}}

\begin{refsection}

\section{Supplementary Method Section}

\subsection{Data collection and annotation}

\begin{figure*}[!htb]
  \centering
  \includegraphics[width=0.9\textwidth]{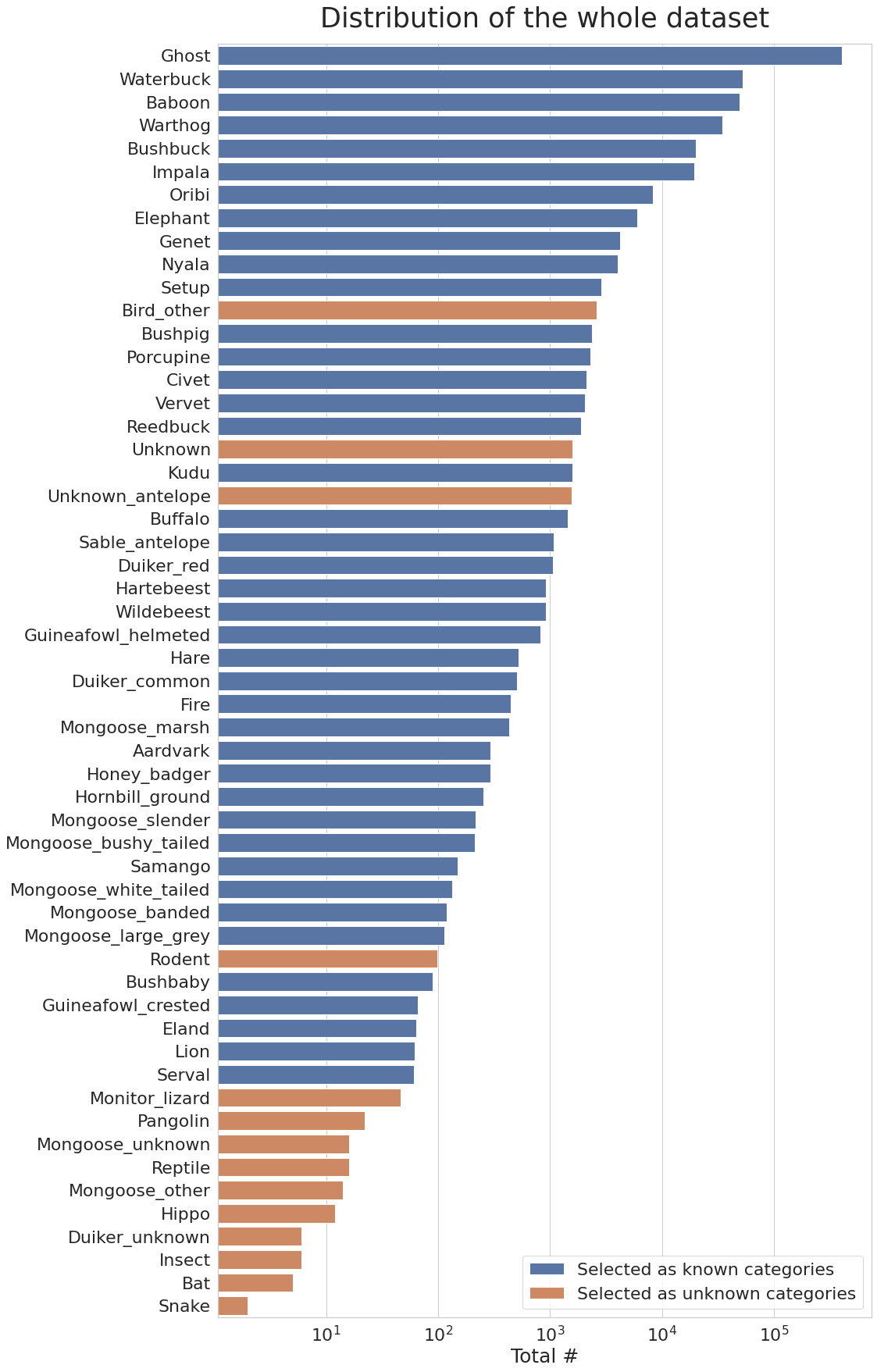}
  \caption{\textbf{The distribution of images across species in the entire camera trap data set.} There are 55 categories in total. 14 categories were tagged as ``unknown'' (colored in orange) and used to improve and validate our model's sensitivity to novel and difficult samples.}
  \label{fig:totaldist}
  \vspace{-10pt}
\end{figure*}

The camera trap data come from the WildCam Gorongosa long-term research and monitoring program in Gorongosa National Park, Mozambique (18.8154\degree S, 34.4963\degree E)~\cite{Gaynor.2020zgf}. The data used in this study are from 2016-2019. Cameras were located in a mix of grassland, open woodland, and closed forest habitats. KMG placed 60 motion-activated Bushnell TrophyCam and Essential E2 cameras in a 300 km2 area in the southern area of the 3,700 km2 park. Each camera was mounted on a tree within 100 meters of the center of a 5 km2 hexagonal grid cell, facing an animal trail or open area with signs of animal activity. Cameras were set in shaded, south-facing sites that were clear of tall grass to reduce false triggers. Cameras took 2 photographs per detection (henceforth, "trigger event") with an interval of 30 seconds between trigger events. There were 630544 images in total. Detailed data distribution with respect to categories is reported in Fig.~\ref{fig:totaldist}. In terms of data split for experimental purposes, the detailed distributions of both group 1 \& 2 are reported in Fig.~\ref{fig:groupdist}.

\subsubsection{Data split}
The data set is randomly split into two groups of training and validation sets to mimic periodical data collection from two sequential time periods, along with an additional ``unknown'' set for improving and validating the model's sensitivity to novel and difficult samples. Because we set the cameras to capture one pair of images for each trigger event, image pairs within the same event are usually similar in appearance. To reduce bias, we split the data set based on camera trigger events, such that both images in a paired trigger events were either both in the training or testing set. The training-testing split did not account for camera locations (i.e., images from a given camera were present in both testing and training sets). For large-scale, long-term projects, it is more likely that the camera locations are stable, and in our study, the cameras cover most of the landscapes in the monitoring area and include a diversity of background types that change seasonally throughout the year.  Possible distribution shifts in our data set solely come from temporal animal community changes instead of spatial landscape/ecosystem changes.

The first group contained the 26 most abundant categories, and the second period contained all 41 categories. We randomly divided each period into training (80\% of samples) and validation (20\% of samples) sets. For scarce categories that had fewer than 80 images (e.g., ``crested guineafowl'', ``eland'', ``lion'', ``serval''), we randomly selected 20 samples instead of 20\% of the data to ensure the quality of validation. The labels and distributions of these two groups of data are illustrated in Fig.~\ref{fig:groupdist} in the Supplementary Method section.

Within the 14 categories that are tagged ``unknown'', we randomly selected 80\% data to fine-tune the model's sensitivity to novel and difficult samples. We then used the rest of the sample from the 14 categories as an extra validation set to evaluate the model's novel image detection capacity.

\subsection{Implementation details}

\begin{figure*}[!htb]
  \centering
  \includegraphics[width=\textwidth]{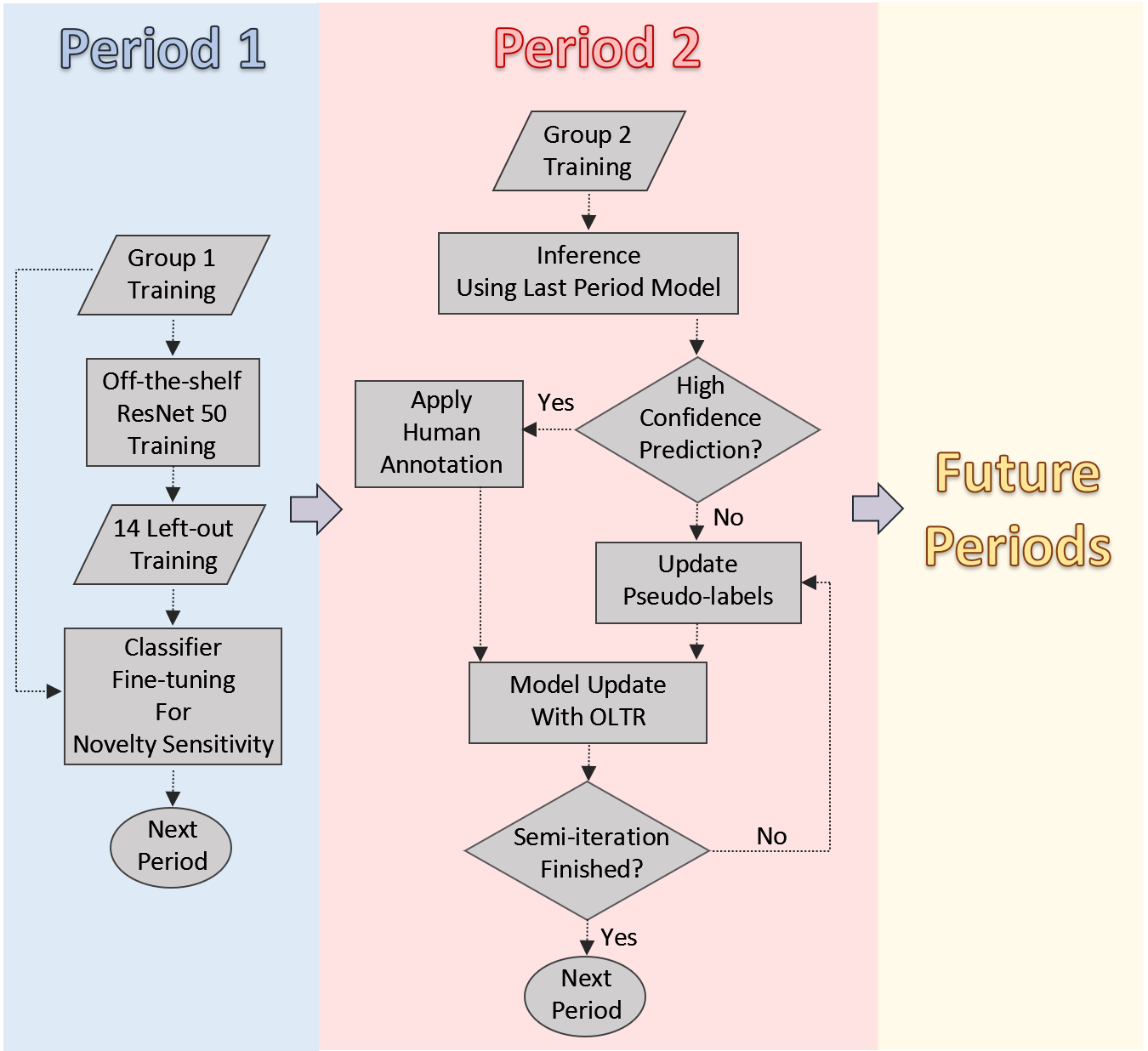}
  \caption{\textbf{The overall experimental workflow of our framework.} In the first time step, a baseline model is trained using group 1 training data with only 26 categories. Next, the classifier is fine-tuned using the 14 unknown categories and energy-based loss to increase the sensitivity to out-of-distribution categories. After the classifier is fine-tuned, the classifier is then used to predict classifications for group 2 training data. Here, high-confidence predictions are trusted while low-confidence predictions are flagged for human annotation. In the final step, both machine- and human-annotations are used to update the previous model with OLTR and semi-supervised techniques. Once the model is updated, the classifier is fine-tuned using energy-based loss again for out-of-distribution sensitivity.}
  \label{fig:workflow}
  \vspace{-10pt}
\end{figure*}

In this section, we report the implementation details of our method. It was developed with Python as the programming language with Pytorch~\cite{pytorch} as the deep learning framework. The detailed experimental pipeline is illustrated in Fig.~\ref{fig:workflow}.

\subsubsection{Data pre-processing}
All of the images used in this project were first resized to $256 \times 256$ dimension. For training inputs, these images were randomly cropped and resized to $224 \times 224$. For validation and inference inputs, images were center cropped to $224 \times 224$. Tab.~\ref{tab:preprocessing} reports the list of data augmentations used for training and corresponding hyper-parameters.

\subsubsection{Period 1 and baseline model training}

There are two steps in this period: 1) baseline model training on group 1 data, and 2) classifier fine-tuning using the 14 ``left-out'' categories for better sensitivity to novel and difficult samples.

\paragraph{Baseline model}
We used ResNet-50~\cite{he2016deep} as our baseline model. It was pre-trained on ImageNet~\cite{deng2009imagenet}, a generalized object oriented data set for model weight initialization. The pre-trained model was then trained on group 1 training data, which has 26 categories. All the hyper parameters can be found in Tab.~\ref{tab:hyperparameters}. Model weights with the best validation performance on group 1 validation data were saved as the best model.

\paragraph{Energy-based fine-tuning}
After training on group 1 data, we used energy-based loss~\cite{liu2020energy} and the 14 ``left-out'' categories (tagged as ``Unknown'') to fine-tune the classifier for better sensitivity to novel and difficult samples. The energy-based loss was calculated as Eq.~\ref{eq:energy_loss}:

\begin{equation}
\begin{split}
\label{eq:energy_loss}
L_{\rm energy} = \mathbb{E}_{x_{\rm known} \sim \mathfrak{D}_{\rm known}^{\rm train}} (\max(0, E(x_{\rm known}) - m_{\rm known})^2 \\ 
+ \mathbb{E}_{x_{\rm unknown} \sim \mathfrak{D}_{\rm unknown}^{\rm train}} (\max(0, m_{\rm unknown} - E(x_{\rm unknown}))^2
\end{split}
\end{equation}

\begin{equation}
\label{eq:energy}
E(x) = -T \cdot \log \sum_i^N e^{(f(x_i) / T)}
\end{equation}
where $\mathbb{E}$ is expectation, $x_{\rm known}$ and $x_{\rm unknown}$ are samples from group 1 and samples from 14 ``Unknown'' categories respectively. $\mathfrak{D}_{\rm known}^{\rm train}$ and $\mathfrak{D}_{\rm unknown}^{\rm train}$ represents data sets of group 1 and 14 ``Unknown'' categories. $E(\cdot)$ is \textit{Helmholtz free energy}, calculated as the log sum of outputs from the network. $f(\cdot): \mathbb{R}^{D \times D} \rightarrow \mathbb{R}^K$ is the network that maps $D \times D$ images to $K$ dimensional vectors. $T$ is temperature that regularize the energy. $m_{\rm known}$ and $m_{\rm unknown}$ are two margins applied on known and unknown energy.

During fine-tuning, both cross-entropy loss and energy-based loss are tuned. Eq.~\ref{eq:final_loss} is the final loss, where $w$ is the weight applied on energy-based loss.

\begin{equation}
\label{eq:final_loss}
L = L_{\rm cross\_entropy} + w \cdot L_{\rm energy}
\end{equation}

All hyper-parameters are reported in Tab.~\ref{tab:hyperparameters}.

\subsubsection{Period 2 and model update}

\paragraph{Active selection and confidence calculation}
Following \cite{liu2020energy}, confidence for active selection is calculated based on \textit{Helmholtz free energy} (Eq.~\ref{eq:energy}). Based on a preset energy threshold $\tau$, predictions are separated into high- and low-confidence. In other words, predictions are considered confident if $-E(x) > \tau$ and vice versa. Based on prediction confidence, low-confidence predictions are assigned human annotations, and high-confidence predictions are utilized as initial pseudo-labels for semi-supervised learning.

\paragraph{Pseudo-labels and semi-supervised learning}
Pseudo-label semi-supervision utilizes both human annotations and pseudo-labels to update the model. In the original approach, where models are randomly initialized, pseudo-labels get updated throughout training iterations~\cite{lee2013pseudo}. In other words, at each iteration, the model predicts samples without human annotations and uses these predictions as pseudo-labels to train the same samples with a stronger set of data augmentations. In our approach, as the pseudo-labels usually have higher quality than random predictions, we set three semi-update repeats and only updated the pseudo-labels in the beginning of each repeat using the best model from last repeat. Specifically, within each semi-update repeat, the model was updated with a fixed set of pseudo-labels and a number of training epochs. Model weights with the best validation performance were saved, and at the end of the repeat, the best model was used to predict samples without human annotations to produce a new set of pseudo-labels, and a new repeat started. Only model weights with the best validation performance throughout the three repeats were saved, and the number of repeats is a hyper-parameter that can be tuned using validation data. Other hyper-parameters can also be found in Tab.~\ref{tab:hyperparameters}.

\paragraph{OLTR}
OLTR is an additional component in our framework targeting the long-tailed distribution of classes in the data sets. Generally speaking, it uses embedding level memory of each category to enhance the distinguishability of scarce categories. It is based on the idea that a lot of the mid-level visual features (i.e., feature embedding) are shared between similar categories (e.g., most of the antelopes share similar body shapes). Since the model can usually learn high quality feature embedding from abundant species, through a memory selection techniques, the model is able to select relevant feature embedding to help improve the distinguishability of scare categories. We directly apply OLTR into our framework. For a detailed explanation of OLTR, please refer to the original paper~\cite{liu2019large}.

\subsection{Comparison to unsupervised and self-supervised learning}
Although unsupervised learning and self-supervised learning have made substantial progress recently~\cite{chen2020simple, he2020momentum} in learning without human annotations, these learning methods still have difficulties handling novel categories and categories with trivial differences (i.e., fine-grained categories)~\cite{xiao2020should}. This is because current unsupervised/self-supervised learning methods rely on human-defined random data augmentation (e.g., cropping and rotation) to mimic intra- and inter-class variations, while real-world novel and fine-grained categories often possess complex intra- and inter-class distributions. In this work, we advocate the use of humans in the loop to provide valuable supervision in a data-efficient manner. Together with semi-supervised learning, our framework can reliably recognize new species with only sparse human annotations.

\subsection{Additional results}
We record detailed results of model update performance by category in Table~\ref{tab:class_perf}.

\subsection{Code availability}

- Code will be released through this GitHub link after submission: \url{https://github.com/zhmiao/AnimalActiveLearning}.

\subsection{Data availability}

All raw camera trap images that were used in this study, along with the associated annotation information, will be uploaded to the publicly-available Labeled Information Library of Alexandria: Biology and Conservation (LILA BC): \url{https://lilablobssc.blob.core.windows.net/gorongosacameratraps/gorongosa-camera-traps-public-256x256.zip}.

\printbibliography[heading=subbibliography]

\clearpage

\clearpage

\begin{figure*}[!htb]
  \centering
  \includegraphics[width=0.9\textwidth]{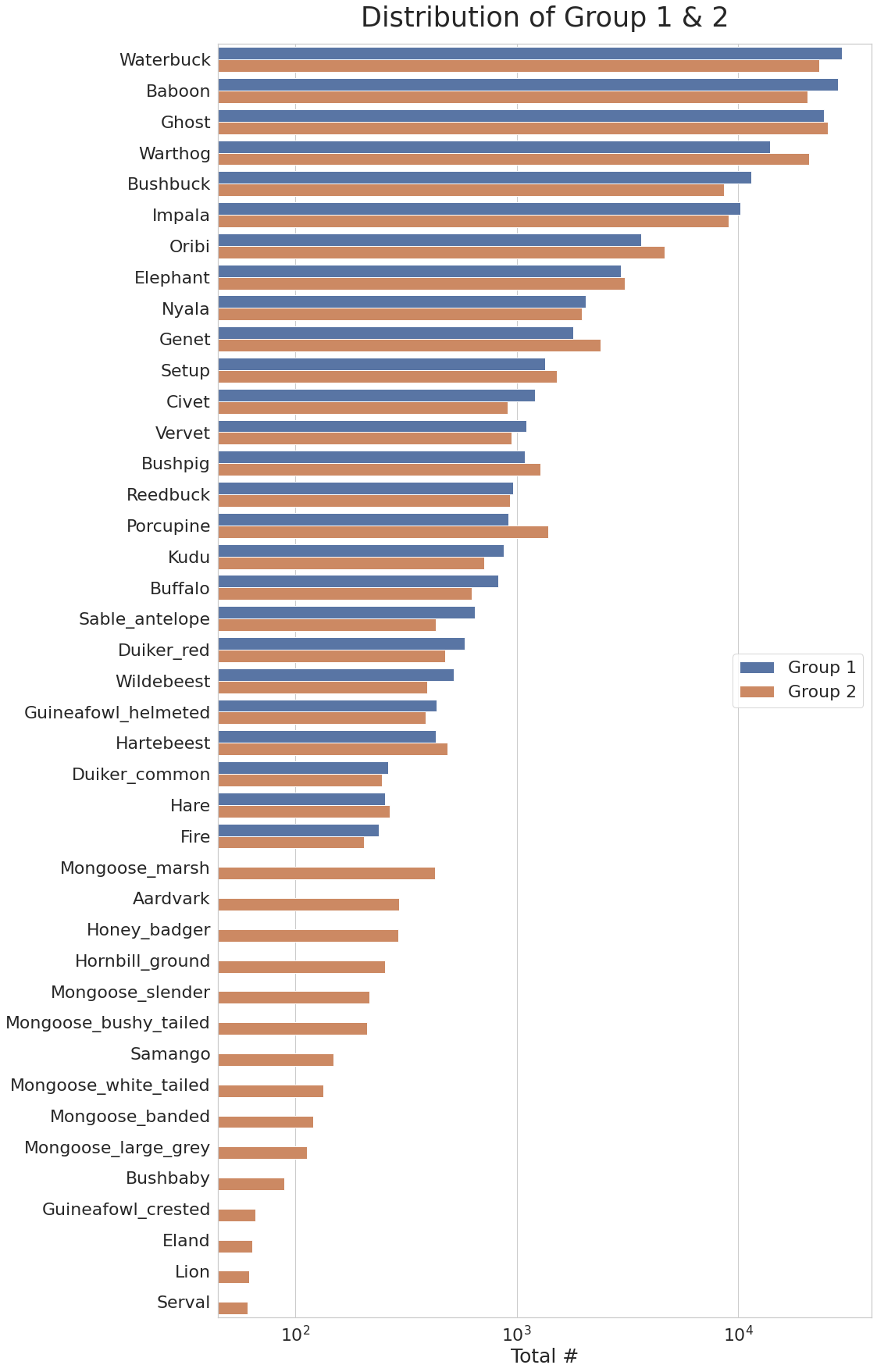}
  \caption{\textbf{The distribution of species across the two groups of data.} We split the data set into two groups to mimic two sequential data collection seasons. In the first group, there are 26 categories (colored in blue). The second group has 41 categories. Group 1 is used in the first period experiment to train a baseline model, and Group 2 is used in the second period experiment to test and update the model.}
  \label{fig:groupdist}
  \vspace{-10pt}
\end{figure*}

\clearpage

\clearpage

\begin{table}[!htb]
\caption{\textbf{List of the augmentation methods and corresponding parameters we used on our training data.}}
\label{tab:preprocessing}
\resizebox{\textwidth}{!}{
\begin{tabular}{
>{\arraybackslash}m{0.3\textwidth} |
>{\arraybackslash}m{0.3\textwidth} 
>{\centering\arraybackslash}m{0.2\textwidth}
}
\hline
Augmentations & Parameters & Values \\
\hline
\multirow{3}{0.35\textwidth}{Random resize crop} 
& Dimension & $224 \times 224$  \\
& Range of crop scale & $0.08 \sim 1.0$ \\
& Range of crop aspect ratio & $0.8 \sim 1.2$ \\
\hline
\multirow{1}{0.35\textwidth}{Random gray scale} 
& Probability & 0.1 \\
\hline
\multirow{1}{0.35\textwidth}{Random horizontal flip} 
& Probability & 0.5 \\
\hline
\multirow{2}{0.35\textwidth}{Random rotation} 
& Probability & 0.5 \\
& Rotation degree & 45 \\
\hline
\multirow{4}{0.35\textwidth}{Color jittering} 
& Brightness jittering & 0.4 \\
& Contrast jittering & 0.4 \\
& Saturation jittering & 0.4 \\
& Hue jittering & 0.1 \\
\hline
\multirow{2}{0.35\textwidth}{Normalization} 
& Mean & $[0.485, 0.456, 0.406]$ \\
& Std & $[0.229, 0.224, 0.225]$ \\
\hline
\end{tabular}}
\end{table}

\clearpage

\begin{table}[!htb]
\caption{\textbf{List of hyperparameters of our framework used in the two-period experiments.}}
\label{tab:hyperparameters}
\resizebox{\textwidth}{!}{
\begin{tabular}{
>{\arraybackslash}m{0.25\textwidth} |
>{\arraybackslash}m{0.4\textwidth} 
>{\centering\arraybackslash}m{0.2\textwidth}
}
\hline
Period & Parameters & Values \\
\hline
\multirow{9}{0.3\textwidth}{Period 1. Training} 
& Baseline architecture &  ResNet-50 \\
& Training epochs & 40 \\
& Batch size & 64 \\
& Initial learning rate (feature) & 0.001 \\
& Initial learning rate (classifier) & 0.01 \\
& Learning rate decay Epochs & 10 \\
& Learning rate decay Ratio & 0.1 \\
& Momentum & 0.9 \\
& Weight decay & 0.0005 \\
\hline
\multirow{9}{0.3\textwidth}{Period 1. Energy Fine-tuning} 
& Training epochs & 10 \\
& Batch size & 96 \\
& Known : Unknown ratio & 1:2 \\
& Energy loss weight & 0.01 \\
& Initial learning rate (feature) & 0.00001 \\
& Initial learning rate (classifier) & 0.0001 \\
& Confidence threshold ($\tau$) & 13.7 \\
& Energy temperature & 1.5 \\
\hline
\multirow{9}{0.3\textwidth}{Period 2. Updating} 
& Baseline architecture &  ResNet-50 + OLTR \\
& Semi-repeats & 3 \\
& Epochs in each repeat & 30 \\
& Batch size & 64 \\
& Pseudo-label \% & 50\% \\
& Initial learning rate of each repeat (feature) & 0.0001 \\
& Initial learning rate of each repeat (classifier) & 0.01 \\
& Initial learning rate of each repeat (memory) & 0.0001 \\
& Learning rate decay Epochs & 10 \\
& Learning rate decay Ratio & 0.1 \\
& Momentum & 0.9 \\
& Weight decay & 0.0005 \\
\hline
\multirow{9}{0.3\textwidth}{Period 2. Energy Fine-tuning} 
& Training epochs & 10 \\
& Batch size & 96 \\
& Known : Unknown ratio & 1:2 \\
& Energy loss weight & 0.01 \\
& Initial learning rate (feature) & 0.000001 \\
& Initial learning rate (classifier) & 0.00001 \\
& Initial learning rate (memory) & 0.000001 \\
& Confidence threshold ($\tau$) & 6.7 \\
& Energy temperature & 0.06 \\
\hline
\end{tabular}}
\end{table}

\clearpage

\begin{table}[!htb]
\caption{\textbf{Classification performance comparisons of Period 2 by category between our method and fully annotated transfer learning.}}
\label{tab:class_perf}
\resizebox{\textwidth}{!}{\begin{threeparttable}
\begin{tabular}{
>{\arraybackslash}m{0.12\textwidth} |
>{\centering\arraybackslash}m{0.2\textwidth}
>{\centering\arraybackslash}m{0.22\textwidth}
>{\centering\arraybackslash}m{0.2\textwidth}
>{\centering\arraybackslash}m{0.22\textwidth}
>{\centering\arraybackslash}m{0.2\textwidth}
}
\hline
& & \multicolumn{2}{m{0.4\textwidth}}{\centering\textbf{Traditional transfer learnig w/ full human ann.}} & \multicolumn{2}{m{0.4\textwidth}}{\centering\textbf{Our framework (Semi-OLTR)}} \\
\hline
& \textbf{Species} & \textbf{\# of Human Ann.} & \textbf{Acc. (\%)} & \textbf{\# of Human Ann.} & \textbf{Acc. (\%)} \\
\hline
\multirow{26}{0.2\textwidth}{Exist in $Group_{1\&2}$} 
& Ghost & 20500 & {\color[HTML]{FE0000} 96.2} & 4248 & 90.2 \\
& Waterbuck & 17938 & {\color[HTML]{FE0000} 88.8} & 2079 & 82.4 \\
& Baboon & 15660 & {\color[HTML]{FE0000} 87.3} & 2335 & 81.1 \\
& Warthog & 17400 & {\color[HTML]{FE0000} 87.4} & 4224 & 79.7 \\
& Bushbuck & 6622 & {\color[HTML]{FE0000} 84.5} & 2179 & 72.3 \\
& Impala & 7153 & {\color[HTML]{FE0000} 84.0} & 1306 & 77.1 \\
& Oribi & 3832 & {\color[HTML]{FE0000} 83.8} & 966 & 76.7 \\
& Elephant & 2471 & {\color[HTML]{FE0000} 88.2} & 470 & 85.1 \\
& Genet & 1976 & {\color[HTML]{FE0000} 85.5} & 888 & 84.0 \\
& Nyala & 1569 & 73.9 & 434 & {\color[HTML]{FE0000} 75.1} \\
& Setup & 1229 & {\color[HTML]{FE0000} 87.4} & 389 & 86.0 \\
& Bushpig & 1040 & 83.1 & 377 & {\color[HTML]{FE0000} 83.1} \\
& Porcupine & 1152 & 83.9 & 300 & {\color[HTML]{FE0000} 88.3} \\
& Civet & 699 & 82.9 & 123 & {\color[HTML]{FE0000} 83.9} \\
& Vervet & 739 & 73.2 & 263 & {\color[HTML]{FE0000} 81.0} \\
& Reedbuck & 740 & 65.8 & 203 & {\color[HTML]{FE0000} 75.3} \\
& Kudu & 556 & 70.9 & 161 & {\color[HTML]{FE0000} 77.2} \\
& Buffalo & 479 & {\color[HTML]{FE0000} 89.0} & 63 & 84.8 \\
& Sable\_antelope & 323 & 85.2 & 48 & {\color[HTML]{FE0000} 86.1} \\
& Duiker\_red & 370 & 86.8 & 116 & {\color[HTML]{FE0000} 89.6} \\
& Hartebeest & 394 & {\color[HTML]{FE0000} 91.2} & 63 & 84.6 \\
& Wildebeest & 303 & {\color[HTML]{FE0000} 83.5} & 44 & 82.4 \\
& Guineafowl\_helmeted & 304 & 64.6 & 250 & {\color[HTML]{FE0000} 74.4} \\
& Hare & 214 & 78.8 & 166 & {\color[HTML]{FE0000} 80.8} \\
& Duiker\_common & 194 & 62.7 & 92 & {\color[HTML]{FE0000} 80.4} \\
& Fire & 160 & 100.0 & 14 & {\color[HTML]{FE0000} 100.0} \\
\hline
\multirow{15}{0.2\textwidth}{Exist in \\ $Group_2$ \\ Only} 
& Mongoose\_marsh & 343 & 70.6 & 287 & {\color[HTML]{FE0000} 71.8} \\
& Aardvark & 235 & 77.6 & 128 & {\color[HTML]{FE0000} 81.0} \\
& Honey\_badger & 234 & 60.3 & 190 & {\color[HTML]{FE0000} 63.8} \\
& Hornbill\_ground & 203 & {\color[HTML]{FE0000} 80.0} & 161 & 72.0 \\
& Mongoose\_slender & 165 & 68.0 & 157 & {\color[HTML]{FE0000} 72.0} \\
& Mongoose\_bushy\_tailed & 161 & {\color[HTML]{FE0000} 74.0} & 106 & 72.0 \\
& Samango & 99 & 58.0 & 48 & {\color[HTML]{FE0000} 70.0} \\
& Mongoose\_white\_tailed & 84 & 52.0 & 79 & {\color[HTML]{FE0000} 64.0} \\
& Mongoose\_banded & 70 & 38.0 & 62 & {\color[HTML]{FE0000} 52.0} \\
& Mongoose\_large\_grey & 63 & 44.0 & 54 & {\color[HTML]{FE0000} 48.0} \\
& Bushbaby & 39 & 36.0 & 31 & {\color[HTML]{FE0000} 50.0} \\
& Guineagowl\_crested & 46 & 95.0 & 35 & {\color[HTML]{FE0000} 100.0} \\
& Eland & 44 & {\color[HTML]{FE0000} 90.0} & 31 & 70.0 \\
& Lion & 42 & 70.0 & 32 & {\color[HTML]{FE0000} 75.0} \\
& Serval & 41 & 45.0 & 32 & {\color[HTML]{FE0000} 60.0} \\
\hline
\end{tabular}
\begin{tablenotes}
\small
\item Red color means higher performance.
\end{tablenotes}
\end{threeparttable}}
\end{table}

\clearpage

%
%

\end{refsection}

\end{document}